\documentclass{article}

\usepackage{microtype}
\usepackage{graphicx}
\usepackage{subcaption}
\usepackage{booktabs} 

\usepackage{hyperref}

\usepackage[preprint]{icml2026}

\usepackage{amsmath}
\usepackage{amssymb}
\usepackage{mathtools}
\usepackage{amsthm}

\usepackage[capitalize,noabbrev]{cleveref}

\theoremstyle{plain}

\theoremstyle{definition}

\theoremstyle{remark}

\usepackage[textsize=tiny]{todonotes}

\icmltitlerunning{Sample Efficient Generative Optimization for Molecular Design}

\begin{document}

\twocolumn[
  \icmltitle{Sample Efficient Generative Optimization for Molecular Design}



  \icmlsetsymbol{equal}{*}

  \begin{icmlauthorlist}
    \icmlauthor{Sarina Kopf}{xxx,yyy,zzz}
    \icmlauthor{Cristina Nevado}{yyy}
    \icmlauthor{Philippe Schwaller}{xxx,zzz}

  \end{icmlauthorlist}

  \icmlaffiliation{yyy}{Department of Chemistry, University of Zurich, Zurich, Switzerland}
  \icmlaffiliation{xxx}{LIAC, ISIC, EPFL, Lausanne, Switzerland}
  \icmlaffiliation{zzz}{Centre of Competence in Research (NCCR) Catalysis, Zurich, Switzerland}

  \icmlcorrespondingauthor{Sarina Kopf}{sarina.kopf@epfl.ch}
  \icmlcorrespondingauthor{Philippe Schwaller}{philippe.schwaller@epfl.ch}

  \icmlkeywords{sample efficiency,generative design,bayesian optimization,molecular design}

  \vskip 0.3in
]



\printAffiliationsAndNotice{}  

\begin{abstract}
Molecular optimization in drug discovery, materials design, and catalysis requires searching vast chemical spaces under tight evaluation budgets, since high-fidelity oracles and experimental measurements are costly. The practical impact of an optimization method therefore hinges on its sample efficiency: how few evaluations it needs to find strong candidates. We introduce Sample Efficient Generative Optimization (SEGO), a framework for Bayesian optimization on adaptively generated molecules. In SEGO, a probabilistic surrogate model forms a hypothesis about where hits lie in chemical space, a generative model is steered to propose candidates in that region, the most promising candidate is selected via an acquisition function, and the resulting oracle call is used both to sharpen the surrogate and to anchor the generator in real reward. SEGO attains state-of-the-art performance on the practical molecular optimization (PMO) benchmark using only one tenth of the oracle calls consumed by other methods, and on a multiparameter docking task it reaches ten hits in roughly half the oracle calls of existing approaches. These gains move molecular optimization closer to campaigns driven by direct experimental feedback.
\end{abstract}
\section{Introduction}
Molecular optimization is a central problem in chemistry, with applications ranging from drug discovery to catalyst design \cite{filella2025optimizing, strieth2024delocalized, seumer2023computational}. These tasks require identifying molecules that satisfy multiple, often competing objectives from vast chemical space \cite{sanchez2018inverse}. Most relevant optimization problems are also data scarce: phenomena such as protein-ligand binding or catalytic activity are difficult to simulate accurately, and experimental measurements---particularly those requiring molecular synthesis---are costly and time-consuming. As a result, the real-world impact of an optimization strategy is defined by the number of evaluations required to discover optimal molecules (sample efficiency). Low sample efficiency necessitates reliance on inexpensive, low-fidelity \textit{in silico} oracles vulnerable to hacking \cite{renz2019failure, lehman2020surprising}. Improved sample efficiency would enable the use of higher-fidelity simulations or even direct experimental feedback, thereby increasing the practical impact of molecular optimization methods. 

ML methods for molecule optimization must traverse chemical space and strategically guide this exploration toward the most promising regions. Traversal methods include deep generative methods that learn a data distribution and can sample from that distribution, including recurrent neural networks (RNNs) \cite{olivecrona-reinvent, reinvent4}, transformers \cite{reinvent-transformer, taiga}, and diffusion or flow-based models \cite{mood, pocketflow, diffsbdd}. These generative methods can be steered with various optimization techniques including Monte Carlo tree search (MCTS), Markov Chain Monte Carlo (MCMC), reinforcement learning (RL), as well as with Bayesian Optimization (BO). Language models that generate molecules in the form of Simplified Molecular Input Line Entry System (SMILES) \cite{smiles} steered with RL remain among the most sample efficient methods \cite{pmo, guo2026sample}. However, state-of-the-art generative methods still require between hundreds and thousands of oracle evaluations to identify high-quality candidates \cite{pmo, guo2026sample}. While this cost may permit the use of moderately expensive simulations such as DFT, it remains prohibitive for workflows involving direct experimental evaluation, particularly in the case of prospective campaigns where the target distribution is not known. 

BO, an optimization technique that uses a probabilistic surrogate model and an acquisition function to guide sequential experimentation within a candidate library or a continuous latent space \cite{frazier2018tutorial}, has excellent sample efficiency and has seen widespread use and success in chemistry \cite{minerva, nanocrystals, doyleboinchem}. Through the fitting of the surrogate model, these methods learn quickly from negative examples. However, performance depends on the presence of suitable candidates within a library or accurate decoding from a latent space to molecule space \cite{agarwal2021discovery, griffiths2022data}. Moreover, BO is highly sensitive to the choice of molecular representation, which can be challenging to design, expensive to compute, and may not generalize across diverse regions of chemical space \cite{rankovic2025large}. Recent BO workflows use large language models (LLMs) trained jointly with GP objectives that enable the rapid, flexible featurization only from SMILES with excellent performance across diverse chemical tasks, opening the door to on-the-fly application across chemical space \cite{kristiadi2024sober, rankovic2025large, rankovic2023bochemian}. 

We propose the Sample Efficient Generative Optimization (SEGO) framework, a tandem system based on RL and BO for sample efficient search in open-ended molecular space. At each iteration, a probabilistic surrogate forms a hypothesis about where hits lie and steers a generative model to produce a library of candidates in that region. The library is featurized on-the-fly through deep-kernel learning over LLM embeddings \cite{rankovic2025large} and an acquisition function selects the most promising candidate for evaluation; the resulting oracle call both sharpens the surrogate into a new hypothesis and anchors the generator in real reward. SEGO achieves sample efficiency from two mechanisms within this loop: SMILES augmentation in the surrogate training, and an anchoring mechanism for the surrogate-guided generation. Our method outperforms both pure BO and pure RL methods and attains state-of-the-art performance on the practical molecular optimization (PMO) benchmark using only a tenth of the oracle budget of other methods. On a multiparameter docking optimization task, SEGO identifies candidates in half or less of the oracle budget of other state-of-the-art methods. These gains could enable prospective campaigns beyond predefined libraries based on direct experimental feedback.

\section{Background and related work}
\subsection{\textit{De Novo} Molecular Design.} \textit{De novo} molecular design seeks to generate molecules with desired properties from scratch, bypassing the constraints of fixed libraries \cite{sanchez2018inverse}. Generative approaches span a broad range of architectures, from SMILES-based recurrent networks and transformers to graph-based, diffusion, and flow-based models. Among these, language-based models operating on SMILES representations have proven to be particularly sample efficient, capable of satisfying even three-dimensional objectives such as docking despite working from one-dimensional string inputs \cite{pmo, guacamol}. A key advantage of SMILES is its non-injective nature: a single molecule admits many valid string representations, and this redundancy can be leveraged through augmentation to improve generalization in low-data regimes \cite{guo2024augmented, arus2019randomized}. As the field has shifted toward stricter oracle budgets that better reflect real-world constraints, sample efficiency has become a central evaluation criterion, and language-based methods consistently rank among the top performers under these conditions. Recent advances such as combining experience replay with SMILES augmentation have pushed sample efficiency further, establishing strong baselines for generative molecular optimization \cite{guo2026sample}.

However, based on the formulation of the loss, these systems do not learn much from negative examples early in the optimization \cite{reinvent4}. In challenging optimization scenarios, where early batches yield little to no reward, slow learning causes long lag times which erode sample efficiency. This motivates integrating generative models with principled selection strategies that extract signal from all evaluations, even from negative evaluations early in the optimization trajectory. 

\subsection{Bayesian Optimization on Molecules.} Bayesian optimization has been applied to molecular discovery across domains in chemistry. For example, it has been used to identify redox-active molecules from libraries of scaffolds enumerated with substituents and featurized with DFT descriptors \cite{agarwal2021discovery}, to identify candidate protein binders from an Enamine library using fingerprint representations \cite{graff2021accelerating}, and for materials discovery efforts, where sets of predefined building blocks were evaluated using time-dependent DFT \cite{strieth2024delocalized}. More broadly, library-based BO has been applied to transition-metal complexes \cite{janet2020accurate}, metal-organic frameworks \cite{comlek2023rapid}, with surrogate models trained on small labeled subsets and acquisition functions used to iteratively select candidates for evaluation. These case studies illustrate the strength of BO for optimization in challenging reward landscapes, but also highlight two recurring limitations: optimization is constrained to expert-crafted libraries that cannot propose structures beyond their initial enumeration, and performance depends heavily on molecular representation. Bespoke descriptors such as DFT features can be expensive to compute, while general-purpose fingerprints, though cheap, may not capture the structure-property relationships relevant to a given task.

Recent work has begun to address the representation challenge through language-model-based molecular embeddings \cite{kristiadi2024sober,rankovic2023bochemian, rankovic2025large}. Building on this direction, the GOLLuM framework fine-tunes language model embeddings using Gaussian Process (GP) marginal likelihood, yielding task-adaptive representations that outperform both fingerprints and bespoke descriptors across diverse benchmarks \cite{rankovic2025large}. Crucially, because GOLLuM operates directly on SMILES strings without requiring additional descriptor computation, it can featurize novel molecules on-the-fly, making it feasible to apply BO over dynamically generated libraries rather than fixed, predefined candidate sets.

\subsection{Hybrid Generative BO methods.} To move beyond fixed libraries, several works have coupled generative models with BO. A common approach embeds discrete molecular representations into a continuous latent space, performs optimization there, and decodes the resulting vectors back into molecules \cite{filella2025optimizing, gomez-vae}. However, this latent-optimize-then-decode paradigm faces practical challenges: decoded molecules are frequently invalid or redundant, and latent spaces can be high-dimensional and poorly structured, complicating surrogate modeling and uncertainty quantification. Alternative approaches bypass latent embeddings and optimize acquisition functions directly over discrete representations using grammar-constrained search \cite{moss2020boss}, graph-based kernels \cite{oh2019combinatorial}, or learned generation policies \cite{swersky2020amortized}, though these methods face their own difficulties in maintaining chemical plausibility and scaling to large batch sizes.

A more recent line of work decouples the generative and optimization stages entirely, first using a generative model to propose candidates and then applying BO to select the most informative subset for evaluation. \citet{dodds2024sample} developed a hybrid framework combining REINVENT with a fingerprint-driven, random-forest-based active learning loop, demonstrating improved sample efficiency over baseline REINVENT, though the modest efficiency of the underlying generative model limits practical applicability. \citet{tripp2024diagnosing} used a genetic algorithm as a molecule generation engine and a Tanimoto kernel on molecular fingerprints to achieve excellent results on the PMO benchmark. \citet{muthyala2025generative} also proposed a generate-then-optimize system with a new acquisition function for batch selection. While SEGO shares this hybrid format, it introduces surrogate-guided generation and augmented optimization to improve sample efficiency substantially. Further, SEGO builds on frontier methods in both Bayesian optimization and generative design, enabling new levels of sample efficiency not achieved by prior hybrid approaches.

\section{SEGO Design}
\begin{figure}[ht]
  \vskip 0.2in
  \begin{center}
    \centerline{\includegraphics[width=\columnwidth]{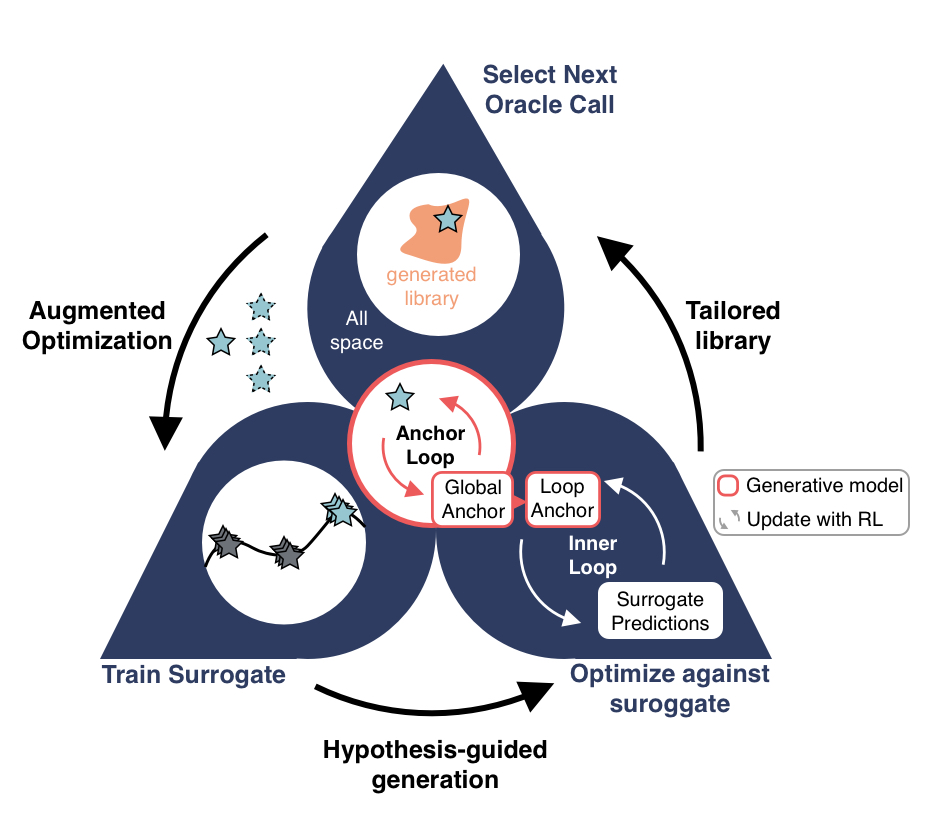}}
    \caption{
Overview of the Sample Efficient Generative Optimization (SEGO) framework. SEGO is a framework for generative optimization that attains excellent sample efficiency through the intelligent allocation of oracle calls across dynamically generated libraries. At each iteration, a surrogate-guided generative model produces a targeted library of molecules in promising regions of chemical space \cite{guo2026sample}. This library is featurized through a deep GP \cite{rankovic2025large}, and the next most promising experiment is selected via an acquisition function. SMILES augmentation during surrogate training and an anchoring mechanism for surrogate-guided generation further improve sample efficiency.
    }
    \label{sego}
  \end{center}
\end{figure}

SEGO is a framework for generative optimization that attains excellent sample efficiency through the intelligent allocation of oracle calls across dynamically generated libraries (Figure \ref{sego}). Each oracle call drives updates to a generative agent, a GP surrogate model, as well as the surrogate's learned featurization. SEGO is organized around the following components: 
\begin{itemize}
    \item An anchor generative agent $\pi_{\theta_{\text{Agent}}}$ updated only on true-oracle feedback.
    \item An inner-loop agent $\pi_{\theta_{\text{Inner}}}$, re-instantiated at each BO iteration from the anchor and trained against the GP posterior mean to produce a dynamic candidate library.
    \item A deep-kernel GP surrogate over LLM embeddings, which featurizes the dynamic library for selection via an acquisition function. 
\end{itemize}
SEGO's sample efficiency rests on two contributions: an anchored surrogate-guided inner loop, enabling surrogate-driven generation while retaining information from true-oracle feedback; and SMILES-augmented surrogate updates, which exploits the string-based featurizer to extract additional signal from each oracle call. 
\subsection{Molecular Generation Engine.} As our generative engine, we take Saturn \cite{guo2026sample}, a Mamba-based \cite{mamba} autoregressive SMILES generator trained with reinforcement learning. A SMILES string $x = (a_1, ..., a_T)$ is generated as a Markov process, 
\begin{equation}
\label{eq:smiles_prob}
P_{\theta}(x) = \prod_{t=1}^{T} \pi_{\theta_{\text{Agent}}}(a_t \mid s_t)
\end{equation}
where $a_t$ is the token selected at step $t$, $s_t$ is the token sequence so far, and  $\pi_{\theta_{\text{Agent}}}$ is the Mamba backbone. The aim in RL is to maximize the expected reward (equation \ref{eq:rl_objective}). 
\begin{equation}
J(\theta) = \mathbb{E}_{a_t \sim \pi_{\theta_{\text{Agent}}}} \left[ \sum_{t=1}^{T} R(a_t, s_t) \right].
\label{eq:rl_objective}
\end{equation}
Augmented likelihood (equation \ref{eq:augmented_likelihood}) is defined, where the prior is a pretrained model with frozen weights, $R$ is a reward function, and $\sigma$ is a scalar modulating the reward's effect:
\begin{equation}
\label{eq:augmented_likelihood}
\log \pi_{\theta_{\text{Augmented}}} = \log \pi_{\theta_{\text{Prior}}} + \sigma R(x)
\end{equation}
Maximizing equation \ref{eq:rl_objective} is equivalent (up to a factor) to minimizing the squared difference between the Augmented and Agent likelihoods (equation \ref{eq:reinvent_loss}). 
\begin{equation}
\label{eq:reinvent_loss}
L(\theta) = \frac{1}{|B|} \left[ \sum_{a \in A^*} \left( \log \pi_{\theta_{\text{Augmented}}} - \log \pi_{\theta_{\text{Agent}}} \right)^2 \right]
\end{equation}

where $B$ is a batch of SMILES sampled from the agent and $A^*$ denotes the action sequences across all timesteps in the batch. Sample efficiency in Saturn comes from Augmented Memory: a replay buffer of the top-100 rewarded SMILES is maintained, randomized into equivalent SMILES, and used for $N_{augmentation}$ additional update rounds per iteration. Mode collapse is mitigated by purging the buffer of scaffolds with low-diversity samples and assigning zero reward to repeated generations beyond a threshold. 
\subsection{Bayesian Optimization Engine.} As our BO engine, we take GOLLuM, a flexible framework based on SMILES that jointly trains LLM embeddings and GP hyperparameters \cite{rankovic2025large}. A learned feature transformation is composed with the base kernel:
\begin{equation}
\label{eq:deep_kernel}
k_{\theta,\phi}(\mathbf{x}, \mathbf{x'}) = k_\theta(g_\phi(\mathbf{x}), g_\phi(\mathbf{x'}))
\end{equation}
where $g_\phi$ is a parameterized feature extractor with parameters $\phi$, trained jointly with the GP hyperparameters through the marginal likelihood. Though multiple ways of constructing $g_\phi$ were proposed in the original paper, we take the projection layer variant: $g_\phi(\mathbf{x}) = \mathbf{P}\text{LLM}(x)$, where $\mathbf{P} \in \mathbb{R}^{m \times d}$ is a learned linear projection followed by ELU activation. The GP hyperparameters $\theta = (\ell, \sigma^2, \sigma_n^2, c)$ and feature parameters $\phi$ are jointly optimized by maximizing the log marginal likelihood:
\begin{align}
\label{eq:marginal_likelihood_gol}
\mathcal{L}(\theta, \phi) &= \log p(\mathbf{y} \mid \mathbf{X}, \theta, \phi) \notag \\
&= -\frac{1}{2} \bigl( \mathbf{y}^\top \mathbf{K}_{\theta,\phi}^{-1} \mathbf{y} + \log |\mathbf{K}_{\theta,\phi}| + n \log 2\pi \bigr)
\end{align}
GOLLuM is particularly well suited as the BO engine in SEGO because it operates directly on SMILES strings, bypassing the need for hand-crafted molecular descriptors. Through joint optimization of the LLM and GP via the marginal likelihood, GOLLuM learns to smooth the representation space so that the GP can generalize effectively from very few observations.

\subsection{SEGO Operation} At iteration $i$, SEGO maintains the anchor agent $\pi_{\theta_{\text{Agent}}}^{(i)}$ and a GP $\mathcal{GP}^{(i)}$ fit to the oracle-evaluated set $\mathcal{D}^{(i)} = \{(x_j, y_j)\}_{j \le i}$. The loop is initialized by sampling $L$ molecules from $\pi_{\theta_\text{Prior}}$, scoring them under the acquisition function, and selecting $n_\text{init}$ for oracle evaluation; these seed $\mathcal{D}^{(0)}$ and provide the first updates to $\pi_{\theta_\text{Agent}}$ and $\mathcal{GP}$. Each iteration:
\begin{enumerate}
    \item Initializes $\pi_{\theta_\text{Inner}}$ from $\pi_{\theta_\text{Agent}}^{(i)}$.
    \item Trains $\pi_{\theta_\text{Inner}}$ for $K_\text{inner}$ Saturn updates against $R_\text{surr}(x) = \mu^{(i)}(x)$, the GP posterior mean.
    \item Samples a candidate library $\mathcal{C}^{(i)}$ of size $L$ from $\pi_{\theta_\text{Inner}}$, then \textit{discards} $\pi_{\theta_\text{Inner}}$.
    \item Selects $x^\star_i = \arg\max_{x \in \mathcal{C}^{(i)}} \alpha(x)$ under acquisition function $\alpha$ (we use expected improvement).
    \item Queries the oracle to obtain $y^\star_i = f_\text{oracle}(x^\star_i)$.
    \item Constructs an augmented set $\mathcal{A}_i = \{(x^\star_{i,k}, y^\star_i)\}_{k=1}^{N_\text{aug}}$ from $N_\text{aug}$ randomized SMILES variants of $x^\star_i$, updates $\mathcal{D}^{(i+1)} \leftarrow \mathcal{D}^{(i)} \cup \mathcal{A}_i$, refits $(\theta, \phi)$ via Eq.~\eqref{eq:marginal_likelihood_gol}, and updates $\pi_{\theta_\text{Agent}}$ by one round of Saturn updates (Eq.~\eqref{eq:reinvent_loss}) with reward $y^\star_i$.
\end{enumerate}

\section{Experiments}
\subsection{SEGO architecture ablation experiments}
We explore the impact of key hyperparameters in SEGO: the role of the inner loop in sample efficiency, the anchor's role in training stability, and augmented optimization. 

\textbf{Experimental Setup.} Following  \citet{guo2024augmented}, we optimize molecules according to a multi-property optimization (MPO) objective: molecular weight (MW) $<$ 350 Da, number of rings $\geq$ 2, and maximize the topological polar surface area (tPSA). To satisfy this objective, molecules must have rings saturated with heteroatoms. Such saturated molecules are dissimilar from training data, so this experiment also evaluates out-of-distribution optimization. We use an extremely stringent oracle budget of 100 calls, representing 10\% of the already stringent budget employed in \cite{guo2026sample}, which better reflects realistic budgets for experimental campaigns. All experiments were run across 10 seeds (0-9 inclusive) using a prior pretrained on the ChEMBL 33 dataset \cite{chembl}. 

\begin{table*}[!ht]
  \caption{Sample efficiency and diversity metrics for inner loop initialization strategies in SEGO. OB (oracle burden) is the number of calls required to generate $N$ unique molecules, and yield is the number of unique molecules generated. IntDiv is internal diversity, Scaffolds is the number of Bemis-Murcko scaffolds, and \#Circles measures chemical space coverage. Metrics are computed at a threshold of 0.7 with a budget of 100 oracle calls. The mean and standard deviation across 10 seeds (0-9 inclusive) are reported. Numbers in parentheses represent number of runs to find any hits out of ten.}
  \label{Inner Loop Initialization}
  \begin{center}
    \begin{small}
      \begin{sc}
        \begin{tabular}{lcccccr}
          \toprule
          Initialization& Yield ($\uparrow$)& OB1($\downarrow$)& OB5($\downarrow$)& IntDiv ($\uparrow$)& Scaffolds ($\uparrow$)& \#Circles ($\uparrow$)\\
          \midrule
          prior& 0$\pm$0&40$\pm$50 (3)&-&-&0$\pm$0&0$\pm$0\\
          continuous& 47$\pm$19&40$\pm$27&38$\pm$17 (9)&0.774$\pm$0.027&47$\pm$19&7$\pm$3\\
          anchor& 57$\pm$22& 23$\pm$6 (9)&30$\pm$7 (9)&0.568$\pm$0.081&52$\pm$21&2$\pm$2\\ \bottomrule
        \end{tabular}
      \end{sc}
    \end{small}
  \end{center}
  \vskip -0.1in
\end{table*}
\textbf{Metrics.} As our sample efficiency metrics, we take \textbf{Yield}, the number of \textit{unique} molecules generated above a reward threshold and \textbf{Oracle Burden} (OB), the number of oracle calls required to generate \textit{N} molecules above a reward threshold. As our threshold, we take 0.7 as molecules start to possess saturated heteroatom rings. We also compute diversity metrics, as there is a tradeoff in most generative methods between yield and diversity. We compute \textbf{IntDiv}, the internal diversity \cite{moses}, \textbf{Scaffolds}, the number of Bemis-Murcko scaffolds \cite{bemis-murcko}, and \textbf{\#Circles}, \cite{circles} which measures chemical space coverage. 

\textbf{Inner loop anchor strategies.} We explore three strategies for instantiating the inner loop, in which the RL agent receives reward from the GP predicted mean. The default strategy initializes each inner loop from an anchor, an RL agent previously trained on true oracle rewards, giving the inner loop a warm start in a promising region of chemical space. To test whether this warm start is necessary, we ablate with two alternatives: a prior strategy, which re-initializes from the pretrained prior at each iteration, testing whether the inner loop can learn to generate high-scoring molecules without any oracle-informed starting point; and a continuous strategy, which retains the inner loop agent across iterations, testing whether it is sufficiently flexible to track the changing GP landscape throughout the campaign.

The prior initialization strategy fails to produce hits in most cases (Table \ref{Inner Loop Initialization}). This is unsurprising: at early iterations the GP has been trained on as few as 10 molecules, so its reward landscape is poorly calibrated across chemical space, providing too faint a signal for the agent to learn from scratch. The continuous strategy yields competitive hit counts, but at the cost of reliability. Since the agent is never reset, it tends to commit to a narrow region of chemical space; if that region contains hits, oracle burden is low, but the agent struggles to transition elsewhere in the absence of hits. This is reflected in the high variance of OB1 across seeds. We adopt the anchored framework as default. 
\begin{table*}[!ht]
  \caption{Sample efficiency and diversity metrics for different augmentation factors of SEGO. OB (oracle burden) is the number of calls required to generate $N$ unique molecules, and yield is the number of unique molecules generated. IntDiv is internal diversity, Scaffolds is the number of Bemis-Murcko scaffolds, and \#Circles measures chemical space coverage. Metrics are computed at a threshold of 0.7 with a budget of 100 oracle calls. The mean and standard deviation across 10 seeds (0-9 inclusive) are reported. Bolded values are statistically significant compared to their counterparts at augmentation factor 1 with a Welch's two-sided t-test at a significance level of 0.01. Numbers in parentheses in the yield column represent number of runs to find any hits out of ten. }
  \label{augmentation_factor}
  \begin{center}
    \begin{small}
      \begin{sc}
        \begin{tabular}{lcccccr}
          \toprule
          Augmentations  & Yield ($\uparrow$)& OB1($\downarrow$)& OB5($\downarrow$)& IntDiv ($\uparrow$)& Scaffolds ($\uparrow$)& \#Circles ($\uparrow$)\\
          \midrule
          1    & 33$\pm$16 &45$\pm$22&61$\pm$17&0.619$\pm$0.063&32$\pm$15&3$\pm$1 \\
          5     & \textbf{65$\pm$6}&25$\pm$7&\textbf{32$\pm$6}&0.601$\pm$0.069&\textbf{62$\pm$8}&3$\pm$2\\
          10    & \textbf{67$\pm$6}& 22$\pm$7&\textbf{29$\pm$6}&0.554$\pm$0.087&\textbf{62$\pm$6}&2$\pm$1  \\
          20    & 55$\pm$21& 25$\pm$10 (9)&\textbf{31$\pm$9} (9)&0.630$\pm$0.070&50$\pm$20&4$\pm$3       \\
          \bottomrule
        \end{tabular}
      \end{sc}
    \end{small}
  \end{center}
  \vskip -0.1in
\end{table*}

\textbf{Surrogate-guided inner loop.} Given the strong performance of the anchor strategy, we next assess the impact of the number of inner loop iterations. To isolate the effect of surrogate-informed training from the additional augmented memory updates that the inner loop provides, we include a further control: a variant with zero inner loops but with \textit{N} anchor augmentation rounds set to 210, matching the total number of updates the 20-step inner loop would receive (20 inner loop steps $\times$ 10 inner loop augmentations, plus 10 outer loop augmentations).
Sample efficiency improves and variance across seeds decreases as the number of inner loop iterations increases, and diversity follows suit, reflecting the stronger optimization (Table \ref{inner loop iters}). The augmentation-matched control, despite receiving the same number of gradient updates from the replay buffer, suffers in yield, oracle burden, and the number of seeds that successfully identify a hit, confirming that the benefit of the inner loop stems from training on GP-predicted rewards rather than from additional augmented memory updates alone. Though the difference between 0 and 20 inner loops does not reach statistical significance for most metrics, the consistent trends across all metrics suggest that the inner loop improves both the reliability and efficiency of SEGO.

\textbf{Augmented Optimization.} Inspired by Augmented Memory, in which a generative model is updated according to $N$ augmentation rounds of randomized, high-reward SMILES, we develop augmented optimization, in which library SMILES are augmented $N$-fold and the GP is trained on all representations for each molecule. Because the GOLLuM framework learns its latent space through implicit contrastive learning — reorganizing so that similar molecules have more similar representations — we hypothesize that providing many representations of the same molecule will encourage the GP to align them, improving predictive accuracy.

Augmented optimization improves both oracle burden and yield without statistically significant impacts on internal diversity and \#Circles (Table \ref{augmentation_factor}). The effect saturates at 5-fold augmentation; no statistically significant difference was observed between 5-, 10-, and 20-fold augmentation. To investigate the mechanism behind this improvement, we examined the impact of augmentation on both representational alignment and predictive accuracy. At each iteration we sampled 500 molecules from the prior and from the anchor and recorded the GP mean prediction across 10 augmentations of each molecule. SMILES augmentation narrows the range of predictions for different representations of the same molecule, and this alignment improves with additional augmentation rounds (Figure \ref{augmentation_range}). Crucially, the effect is observed for the prior as well as the anchor, indicating that augmentation improves global representational alignment rather than only aligning the narrow region of chemical space being explored by the oracle. This corresponds to gains in predictive accuracy — measured by the correlation between mean predicted values and true oracle values — particularly early on, when the GP has very little training data (Figure \ref{augmentation_range}).

\textbf{Isomer Collapse.} The previous experiments showcase the powerful optimization behavior of SEGO, though as is typical of strong optimizers, this comes at the cost of diversity. The Bemis-Murcko scaffold metric even inflates diversity for this oracle: dense saturation means that molecules differing in only heteroatom placement are assigned different scaffolds. To probe diversity collapse for this oracle, we introduce the metric \textbf{Isomers}, defined as the number of unique hit molecular formulas. Despite generating 59 ($\pm$25) unique molecules at this threshold, SEGO produces only 12 ($\pm$9) unique isomers, revealing a form of mode collapse in which the agent repeatedly generates constitutional isomers of the same formula (Table \ref{tab:eixi}).

We address this isomer collapse by modulating the expected improvement acquisition function. Standard expected improvement can favor isomers predicted to score near the previous best over novel formulas whose higher uncertainty does not translate into a higher chance of exceeding the incumbent. Following \citet{brochu}, we inflate the incumbent by a margin $\xi$ and so require a candidate's predicted improvement to clear this raised threshold. We observe that $\xi=0.05$ and $\xi=0.1$ significantly increase the number of unique hit isomers, with no decrease in the overall number of hits (Table \ref{tab:eixi}, implementation details in section \ref{app:eixi}). We also introduce an isomer filter analogous to a Bemis-Murcko scaffold filter, setting a stringent threshold of 2. This filter offers additional gains when combined with the inflated expected improvement, showing significant improvements over the baseline in isomers and both diversity metrics. Both the $\xi$ parameter and the isomer bucket size are hyperparameters that can be tuned to balance exploration and exploitation. 
\begin{table*}[!bt]
  \caption{Sample efficiency and diversity metrics comparing SEGO to its engines, Saturn (generative backbone) and GOLLuM (BO). OB (oracle burden) is the number of calls required to generate $N$ unique molecules, and yield is the number of unique molecules generated. IntDiv is internal diversity, Scaffolds is the number of Bemis-Murcko scaffolds, and \#Circles measures chemical space coverage. Metrics are computed at a threshold of 0.7 with a budget of 100 oracle calls. The mean and standard deviation across 10 seeds (0-9 inclusive) are reported. Numbers in parentheses in the yield column represent number of runs to find any hits out of ten. }
  \begin{center}
    \begin{small}
      \begin{sc}
        \begin{tabular}{lcccccr}
          \toprule
          Method& Yield ($\uparrow$)& OB1($\downarrow$)& OB5($\downarrow$)& IntDiv ($\uparrow$)& Scaffolds ($\uparrow$)&  \#Circles ($\uparrow$)\\\midrule
 ChEMBL 33& 0$\pm$0& -& -& -& 0$\pm$0&0$\pm$0\\
          
          Saturn& 0$\pm$0&-&-&-&0$\pm$0&0$\pm$0\\
          GOLLuM-1x& 0$\pm$1&18$\pm$1 (2)&-&-&0$\pm$1&0$\pm$1\\
 GOLLuM-10x& 0$\pm$0& 60$\pm$29 (2)& -& -& 0$\pm$0&0$\pm$0\\
          SEGO& 63$\pm$9& 24$\pm$8&32$\pm$7&0.702$\pm$0.079&59$\pm$10&6$\pm$3\\ \bottomrule
        \end{tabular}
      \end{sc}
    \end{small}
  \end{center}
  \label{tab:toyres}
  \vskip -0.1in
\end{table*}

\textbf{Hybrid versus individual approaches.} SEGO is a hybrid architecture that enables sample-efficient search over open-ended chemical space. We compare SEGO to the individual generative and BO engines it is built from to showcase the benefits of the hybrid approach. To benchmark GOLLuM, we take a random sample of 500 molecules from ChEMBL 33 \cite{chembl}, the same database used to train the prior in Saturn and in SEGO, to mimic a real-world prospective campaign in which a desirable library is not known. Given the benefits of augmented optimization observed earlier, we also test GOLLuM with 10-fold SMILES augmentation (GOLLuM-10x). For SEGO, we use the anchored framework with 20 inner loop iterations and an augmented optimization factor of 10. We also use $\xi=0.05$ as well as an Isomer diversity filter with a bucket size of 2. 

Saturn fails to produce any hits at this strict oracle budget (Table \ref{tab:toyres}). With the recommended batch size of 16 and augmentation rounds of 10, it undergoes only 6 rounds of RL updates, which is insufficient to shift the distribution toward hit molecules. Smaller batch sizes in Saturn do not lead to gains in sample efficiency due to the noise in approximating expected reward with small batch sizes \cite{guo2026sample}. GOLLuM fails because the randomly sampled library does not contain hits. In SEGO, the surrogate-steered generation shifts the distribution toward high reward regions providing the selector with hit-containing libraries to yield 63 ($\pm9$) hits.

\subsection{PMO benchmark}
\textbf{Experimental Setup.} We evaluate SEGO on the practical molecular optimization (PMO) benchmark, a suite of 23 oracles spanning model-based scoring (e.g., DRD2, JNK3, GSK3$\beta$), similarity-based objectives (e.g., Albuterol, Mestranol, Celecoxib rediscovery), and physicochemical and multi-property composites (e.g., QED, isomer matching, scaffold hopping) \cite{pmo}. PMO reports the sum of each trajectory's area under the curve (AUC) of the top-10 average performance as a function of cumulative oracle queries, favoring methods that find high values in fewer calls. The PMO takes a relaxed budget of 10,000. For increased relevance to expensive oracles, following \citet{lico}, we instead report results for baselines on the PMO when restricted to an oracle budget of 1,000 (PMO-1k). For SEGO, we restrict oracle budget to just 100 oracle calls, after which the top-1 and top-10 scores are held fixed for computing metrics at the 1k horizon \cite{seismo} to showcase sample efficiency. As many objectives are similarity based, and therefore diversity is not a priority, we run SEGO with vanilla expected improvement as the acquisition function. We report both top-1 AUC (consistent with 1\% of oracle calls) and top-10 AUC, aggregated across 3 seeds (0--2 inclusive). We compare against published PMO leaderboard methods including REINVENT \cite{reinvent4}, Augmented Memory \cite{guo2024augmented}, Genetic GFN \cite{geneticgfn}, MOLLEO \cite{molleo}, Graph GA \cite{graphga}, and GP BO \cite{tripp2024diagnosing}.

\begin{table}[!b]
  \caption{Performance of SEGO and baselines on 23 tasks in the PMO-1k. A higher score is better. We report the mean and standard deviation of 3 seeds. Baseline results are taken from prior work \cite{lico} for the top-10 hits (1\% of oracle budget). SEGO only proposes molecules during the first 100 oracle calls, after which the top-1 (top 1\%) or top-10 (top 10\%) score is held fixed, while baselines continue optimizing for the full 1,000 oracle calls.} 
  \label{pmo}
  \begin{center}
    \begin{small}
      \begin{sc}
        \begin{tabular}{lc}\toprule
          
          Method& Sum of AUC ($\uparrow$)\\\midrule
 REINVENT& 10.68\\
          
          Augmented Memory& 10.81\\
          Graph GA& 10.90\\
 GP BO& 11.27\\
          Genetic GFN& 11.56\\ 
 MOLLEO& 11.65\\
 LICO&11.71\\
 SEGO-top10 (ours)& 11.16\\
 SEGO-top1 (ours)& 11.86\\ \bottomrule
        \end{tabular}
      \end{sc}
    \end{small}
  \end{center}
  \vskip -0.1in
\end{table}

\begin{table*}[!ht]
  \caption{Oracle Burden@10 comparisons between GEAM, Saturn, and SEGO. \textbf{OB@10} is the number of calls required to generate 10 unique strict hit molecules (lower is better). The mean and standard deviation across 10 seeds (0-9 inclusive) are reported. Numbers in parentheses represent successful seeds.}
  \label{tab:geam}
  \begin{center}
    \begin{small}
      \begin{sc}
        \begin{tabular}{lccccc}
          \toprule
          Method& parp1& fa7& 5ht1b& braf& jak2\\
          \midrule
          GEAM& 743$\pm$52&1,446$\pm$404&531$\pm$38&892$\pm$144&537$\pm$70\\
          Saturn& 518$\pm$92&924$\pm$247&105$\pm$23&581$\pm$123&348$\pm$96\\ 
 SEGO (ours)& 133$\pm$127& 244$\pm$113 (9)& 53$\pm$12& 208$\pm$122& 90$\pm$21\\ 
 \bottomrule
        \end{tabular}
      \end{sc}
    \end{small}
  \end{center}
  \vskip -0.1in
\end{table*}

At the top-1 threshold, which mirrors the baselines by evaluating the top 1\% of the oracle budget, SEGO achieves state-of-the-art performance \textbf{despite consuming }\textbf{only a tenth of the oracle budget available to other methods} (Table \ref{pmo}). SEGO attains competitive performance at the top-10 threshold which is particularly challenging because top-10 represents 10\% of the total budget available to SEGO. Extended results for individual oracles are reported in the appendix (Table \ref{pmo-detailed}). Together, these results demonstrate SEGO's strong optimization capability under tight budget constraints. 

The most closely related method in this benchmark is GP BO \cite{tripp2024diagnosing}, which uses BO to select candidates from a library generated by a genetic algorithm. SEGO outperforms GP BO at a fraction of the oracle budget, highlighting the strength of its underlying generative and optimization components. LICO \cite{lico} follows a similar framework but replaces the GP surrogate with an LLM. SEGO surpasses LICO with greater sample efficiency, though LICO's strong results suggest that substituting SEGO's GP with an LLM is a promising direction for future work. Another language-model approach is MOLLEO \cite{molleo}, which leverages an LLM to perform mutation and crossover operations within an evolutionary algorithm. While MOLLEO achieves competitive performance, it relies on detailed textual task descriptions provided to the LLM, raising concerns about data leakage and limiting its applicability to prospective campaigns where such descriptions may not be available.

\subsection{Molecular Docking Task}
\textbf{Experimental Setup.} Following \citet{geam}, we benchmark SEGO on multiparameter optimization docking tasks against five targets (parp1, fa7, 5ht1b, braf and jak2). We optimize the MPO objective function (equation \ref{eq:geamscore}), where $\widehat{\mathrm{DS}}$ is the normalized QuickVina 2 \cite{quickvina2} score and $\widehat{\mathrm{SA}}$ is the normalized synthetic accessibility score. We compare against the goal-aware fragment extraction, assembly and modification (GEAM) model \cite{geam} and against Saturn. 
\begin{equation}
R(\mathbf{x}) = \widehat{\mathrm{DS}}(\mathbf{x}) \times \mathrm{QED}(\mathbf{x}) \times \widehat{\mathrm{SA}}(\mathbf{x}) \in [0, 1]
\label{eq:geamscore}
\end{equation}
\textbf{Metrics.} Mirroring a campaign in which the aim is to find several promising candidates for further evaluation and in which experiments are expensive, we take the oracle burden at 10 hits (\textbf{OB@10}). Here, we define a hit as a strict hit with stringent criteria requiring a molecule to have a docking score better than the median of known actives, a $QED>0.7$ and $SA<3$ \cite{guo2026sample}. We take strict hits to test the optimization capacity of the model. SEGO was run for an oracle budget of 600 calls, Saturn and GEAM were run for 3,000 calls. 

\textbf{Results.} SEGO finds 10 strict hits in half or less of the oracles required by Saturn and GEAM, showcasing strong optimization performance on a pharmacologically relevant oracle (Table \ref{tab:geam}). Generated molecules for each target are shown in Figure \ref{fig:samplemols}. 

\section{Discussion}
We introduce SEGO, a hybrid generative optimizer for sample-efficient search in open-ended chemical space. Unlike pure generative methods that rely on slow updates to shift the distribution and conventional BO methods that operate on fixed libraries, SEGO leverages the principled exploration of BO across dynamically generated libraries. This design yields gains in hit-finding and sample efficiency compared to the individual parent methods on a toy oracle. SEGO also achieves excellent performance on the PMO benchmark, reaching state-of-the-art top 1\% performance in only 100 oracle calls---one-tenth of the oracle budget of PMO-1k. On a more challenging MPO molecular docking oracle, SEGO identifies multiple hits much faster than current state-of-the-art methods.

Our ablations show that our design decisions—namely, guiding an inner loop based on surrogate predictions and anchoring that inner loop—contribute to the sample efficiency of SEGO. Further, data augmentation, known to be beneficial for sample efficiency in generative models, can improve the predictive performance of BO as well and improve representational alignment between diverse SMILES. Together, this setup yields an effective optimizer that is prone to repeat isomer generation. We enhance diversity through the inclusion of an isomer-based diversity filter as well as by inflating the previous best in the expected improvement acquisition function. 

Excellent performance on the PMO benchmark and the MPO docking benchmark hint at powerful optimization ability; however, it may not reflect optimization behavior in the complex reward landscapes found in real-world campaigns where competing objectives like synthetic accessibility, solubility, and activity must be navigated simultaneously. Further work should assess SEGO on challenging oracles including physics-based simulations relevant to drug-discovery and catalysis as well as stringent synthesis constraints before considering deployment to wet-lab prospective campaigns \cite{lee2024molecule}. Finally, practical high-throughput campaigns require multiobjective optimization over Pareto fronts and batched acquisition \cite{minerva}  which could be introduced into SEGO via alternative acquisition functions \cite{muthyala2025generative}. Despite these limitations, the sample efficiency and hit-finding ability demonstrated here are a promising step toward generative molecular design driven by direct experimental feedback. 

\section*{Software and Data}
Software to run SEGO can be found at \url{https://github.com/schwallergroup/sego}.
\section*{Impact Statement}
As AI-driven methods become increasingly integrated into chemical discovery, it is important to acknowledge the dual-use risks inherent in molecular generation, which become more relevant as engines become more powerful. The risks of SEGO are in line with existing performant molecular design engines. Such risks can be mitigated through controlled access to oracles and established safeguards at the institutional level. 

\section*{Acknowledgments}
This publication was created as part of NCCR Catalysis (grant number 225147), a National Centre of Competence in Research funded by the Swiss National Science Foundation. 


\bibliography{sego}
\bibliographystyle{icml2026}


\newpage
\appendix
\onecolumn
\setcounter{figure}{0}
\setcounter{table}{0}
\renewcommand{\thefigure}{A\arabic{figure}}
\renewcommand{\thetable}{A\arabic{table}}
\section{Implementation details.}
\subsection{Saturn Hyperparameters.} Saturn was used with default hyperparameters as recommended in the paper \cite{guo2026sample}. The inner loop used 10 augmentation rounds and a batch size of 64, selected to be a good balance between sample efficiency and stability in the surrogate-guided approach which does not consume oracle calls. The outer loop used 10 augmentation rounds based on a batch size of 1, selected by the BO loop. Unless otherwise stated, the ablation studies use only a Bemis-Murcko scaffold diversity filter. The toy-oracle hybrid comparison and the benchmark studies additionally include an isomer diversity filter in both the inner loop and the outer loop, with a default bucket size of 2.  

\subsection{GOLLuM Hyperparameters.} GOLLuM was used with default hyperparameters as recommended in the paper \cite{rankovic2025large}. We used the T5 language model \cite{T5} as our base featurizer in the PLLM setting, using a Gaussian process \cite{seeger2004gaussian} as our surrogate model and a Matérn-5/2 kernel. We use the expected improvement acquisition function. For ablation studies and the PMO benchmark, we used a $\xi$ value of 0.0; for the docking benchmark and the toy oracle we used a $\xi$ value of 0.05. For other implementation details, refer to \citet{rankovic2025large}.

\subsection{Expected Improvement Modulation.}
\label{app:eixi}
Following \citet{brochu}, we adjust the balance between exploration and exploitation by tuning the incumbent $f(\mathbf{x}^{+}) $ by $\xi$ (equations \ref{eq:ei}, \ref{eq:ei-z}).
\begin{equation}
\mathrm{EI}(\mathbf{x}) =
\begin{cases}
\bigl(\mu(\mathbf{x}) - f(\mathbf{x}^{+}) - \xi\bigr)\,\Phi(Z)
  + \sigma(\mathbf{x})\,\phi(Z) & \text{if } \sigma(\mathbf{x}) > 0, \\[4pt]
0 & \text{if } \sigma(\mathbf{x}) = 0,
\end{cases}
\label{eq:ei}
\end{equation}

\begin{equation}
Z =
\begin{cases}
\dfrac{\mu(\mathbf{x}) - f(\mathbf{x}^{+}) - \xi}{\sigma(\mathbf{x})}
  & \text{if } \sigma(\mathbf{x}) > 0, \\
0 & \text{if } \sigma(\mathbf{x}) = 0.
\end{cases}
\label{eq:ei-z}
\end{equation}

\section{Extended Ablation Results}
\subsection{Inner Loop Iterations}

\begin{table*}[!ht]
  \caption{Sample efficiency and diversity metrics for inner loop iterations. The first entry sets the anchor augmentation rounds to 210 to mirror the number of updates that would occur in the inner loop given augmented memory. OB (oracle burden) is the number of calls required to generate $N$ unique molecules, and yield is the number of unique molecules generated. IntDiv is internal diversity, Scaffolds is the number of Bemis-Murcko scaffolds, and \#Circles measures chemical space coverage. Metrics are computed at a threshold of 0.7 with a budget of 100 oracle calls. The mean and standard deviation across 10 seeds (0-9 inclusive) are reported. Numbers in parentheses represent number of runs to find any hits out of ten. }
  \label{inner loop iters}
  \begin{center}
    \begin{small}
      \begin{sc}
        \begin{tabular}{lcccccr}
          \toprule
          Inner Loop Iterations& Yield ($\uparrow$)& OB1($\downarrow$)& OB5($\downarrow$)& IntDiv ($\uparrow$)& Scaffolds ($\uparrow$)& \#Circles ($\uparrow$)\\
          \midrule
          0 match augmentation& 24$\pm$31&34$\pm$25 (4)&42$\pm$25 (3)&0.423$\pm$0.052&12$\pm$17&1$\pm$1\\
          0& 51$\pm$24&40$\pm$29&39$\pm$21 (9)&0.519$\pm$0.084&37$\pm$23&1$\pm$1\\
          5& 50$\pm$25& 33$\pm$18 (9)&39$\pm$18 (9)&0.580$\pm$0.087&43$\pm$22&2$\pm$2\\
 10& 60$\pm$14& 31$\pm$15 & 37$\pm$16& 0.582$\pm$0.092& 54$\pm$13&3$\pm$1\\
 20& 66$\pm$9& 23$\pm$9& 30$\pm$9& 0.608$\pm$0.086& 60$\pm$9&4$\pm$2\\ \bottomrule
        \end{tabular}
      \end{sc}
    \end{small}
  \end{center}
  \vskip -0.1in
\end{table*}
\newpage
\subsection{SMILES augmentation}
\begin{figure}[ht]
  \vskip 0.2in
  \begin{center}
    \centerline{\includegraphics[width=\columnwidth]{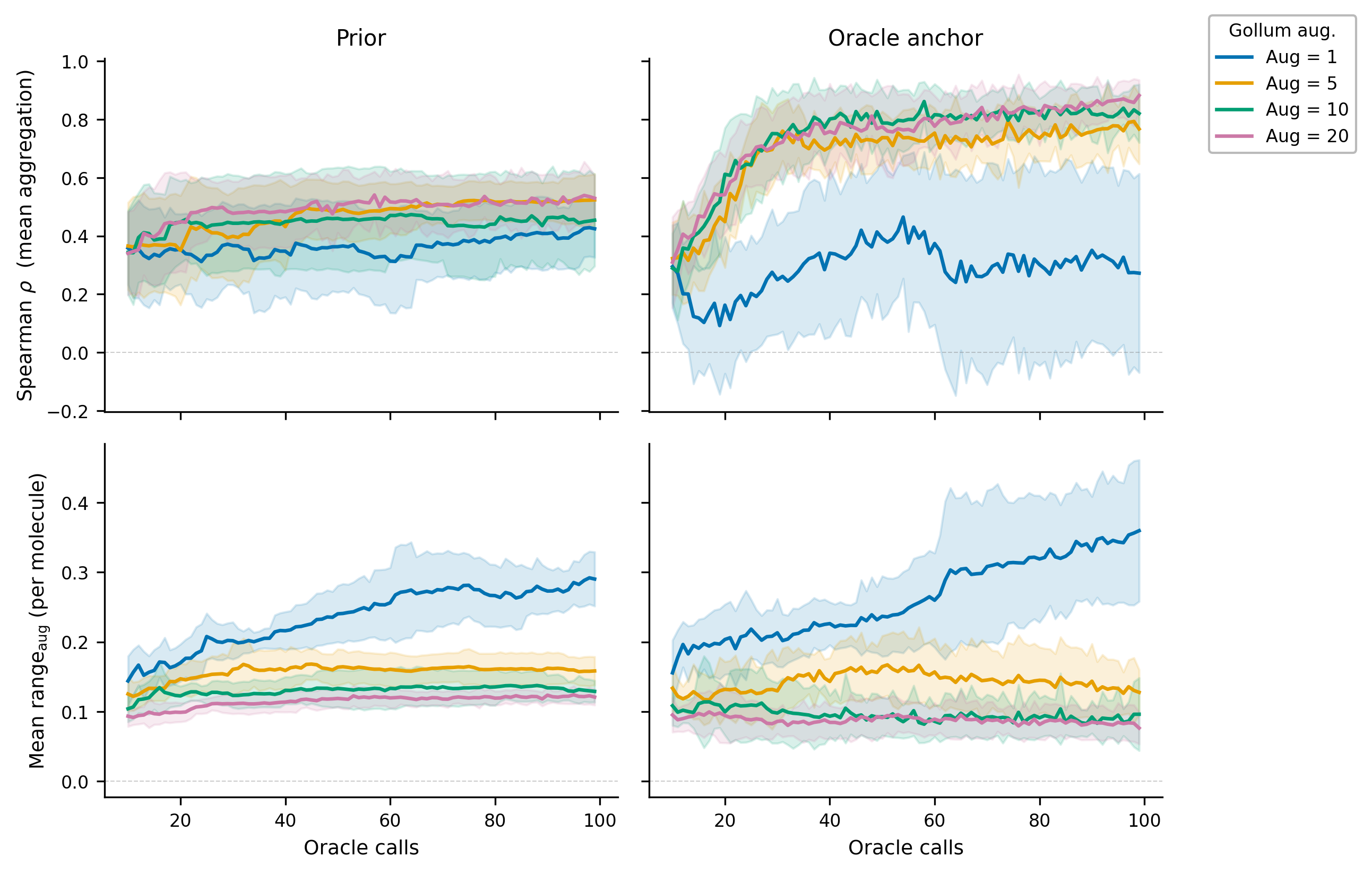}}
    \caption{Effect of SMILES augmentation on GP surrogate quality (predictive
    accuracy and representational alignment) for molecules sampled from the prior
    and the oracle anchor across oracle calls. \textbf{Top:} Spearman rank
    correlation ($\rho$) between the GP's predicted rewards (mean over 10 augmented
    SMILES representations) and true oracle rewards, averaged over 10 seeds
    (shading: $\pm 1$ std). \textbf{Bottom:} Mean within-molecule prediction range
    (max $-$ min GP prediction across 10 augmentations per molecule), measuring
    consistency of the GP featurizer. Without augmentation during training
    (aug~$=1$), the GP achieves poor rank correlation on oracle-anchor molecules and
    produces highly inconsistent predictions across SMILES representations of the
    same molecule. Increasing to aug~$\geq5$ substantially reduces within-molecule
    variance and improves rank correlation, particularly on the oracle-anchor set.
    Further increases to aug~$=10$ or $20$ yield diminishing returns. Notably,
    augmentation improves representational alignment even for molecules sampled from
    the prior---diverse chemical space far from the oracle-explored region---though
    the gains in predictive accuracy there are more modest.}
    \label{augmentation_range}
  \end{center}
  \end{figure}
\newpage
  \subsection{Diversity}
  
  \begin{table*}[!ht]
  \caption{Sample efficiency and diversity metrics for SEGO with a $\xi$ expected improvement factor and with an isomer diversity filter. OB (oracle burden) is the number of calls required to generate $N$ unique molecules, and yield is the number of unique molecules generated. IntDiv is internal diversity, Scaffolds is the number of Bemis-Murcko scaffolds, and \#Circles measures chemical space coverage. Metrics are computed at a threshold of 0.7 with a budget of 100 oracle calls. The mean and standard deviation across 10 seeds (0-9 inclusive) are reported. Bolded values are statistically significant compared to the baseline with a Welch's two-sided t-test at a significance level of 0.05. Numbers in parentheses in the yield column represent number of runs to find any hits out of ten.}
  \begin{center}
  \footnotesize
      \begin{sc}
        \begin{tabular}{lccccccr}
          \toprule
          Diversity Filter& Yield ($\uparrow$)& OB1($\downarrow$)& OB5($\downarrow$)& IntDiv ($\uparrow$)& Scaffolds ($\uparrow$) &Isomers ($\uparrow$)& \#Circles ($\uparrow$)\\
          \midrule
          $\xi=0.0$& 59$\pm$25&28$\pm$18 (9)&35$\pm$16 (9)&0.576$\pm$0.096&54$\pm$23&12$\pm$9&2$\pm$1\\
          $\xi=0.05$& 68$\pm$8&25$\pm$9 &31$\pm$9&0.644$\pm$0.045&63$\pm$10&\textbf{23$\pm$7}&3$\pm$1\\ 
 $\xi=0.1$& 58$\pm$20& 26$\pm$10& 39$\pm$23& 0.672$\pm$0.073& 57$\pm$20& \textbf{28$\pm$12}&\textbf{4$\pm$2}\\
 $\xi=0.05$ + Isomer Filter& 63$\pm$9& 24$\pm$8& 32$\pm$7& \textbf{0.702$\pm$0.079}& 59$\pm$10& \textbf{46$\pm$14}&\textbf{6$\pm$3}\\
 $\xi=0.1$ + Isomer Filter& 60$\pm$9& 28$\pm$10& 36$\pm$11& \textbf{0.712$\pm$0.036}& 58$\pm$9& \textbf{52$\pm$6}&\textbf{7$\pm$3}\\
 \bottomrule
        \end{tabular}
      \end{sc}
  \end{center}
  \label{tab:eixi}
  \vskip -0.1in
\end{table*}

\section{Extended Benchmark Results}

\begin{table*}[!ht]
  \caption{Performance of SEGO and baselines on 23 tasks in the PMO-1k. A higher score is better. We report the mean and standard deviation of 3 seeds. SEGO only proposes molecules during the first 100 oracle calls, after which the top-1 or top-10 score is held fixed. We report the top-1 and top-10 AUC score across 100 oracle calls, 1k oracle calls, and 10k oracle calls.}
  \label{pmo-detailed}
  \begin{center}
    \begin{small}
      \begin{sc}
\begin{tabular}{lcccccc}
\toprule
 Budget& \multicolumn{2}{c}{100} & \multicolumn{2}{c}{1k} & \multicolumn{2}{c}{10k} \\
Oracle& top-1 & top-10 & top-1 & top-10 & top-1 & top-10 \\
\midrule
albuterol\_similarity& 0.492$\pm$0.114 & 0.437$\pm$0.076 & 0.568$\pm$0.139 & 0.537$\pm$0.113 & 0.576$\pm$0.142 & 0.547$\pm$0.117 \\
amlodipine\_mpo& 0.475$\pm$0.027 & 0.429$\pm$0.028 & 0.524$\pm$0.040 & 0.509$\pm$0.036 & 0.529$\pm$0.041 & 0.517$\pm$0.037 \\
celecoxib\_rediscovery& 0.350$\pm$0.032 & 0.312$\pm$0.026 & 0.405$\pm$0.051 & 0.385$\pm$0.045 & 0.410$\pm$0.053 & 0.392$\pm$0.047 \\
deco\_hop& 0.592$\pm$0.012 & 0.582$\pm$0.013 & 0.605$\pm$0.013 & 0.602$\pm$0.014 & 0.606$\pm$0.013 & 0.604$\pm$0.014 \\
drd2 & 0.400$\pm$0.383 & 0.315$\pm$0.376 & 0.718$\pm$0.317 & 0.605$\pm$0.413 & 0.749$\pm$0.315 & 0.634$\pm$0.423 \\
fexofenadine\_mpo& 0.647$\pm$0.024 & 0.597$\pm$0.026 & 0.717$\pm$0.017 & 0.703$\pm$0.019 & 0.724$\pm$0.017 & 0.713$\pm$0.019 \\
gsk3b & 0.390$\pm$0.065 & 0.309$\pm$0.086 & 0.603$\pm$0.047 & 0.577$\pm$0.029 & 0.624$\pm$0.056 & 0.604$\pm$0.039 \\
isomers\_c7h8n2o2& 0.713$\pm$0.077 & 0.586$\pm$0.089 & 0.956$\pm$0.020 & 0.896$\pm$0.017 & 0.981$\pm$0.015 & 0.927$\pm$0.010 \\
isomers\_c9h10n2o2pf2cl& 0.463$\pm$0.122 & 0.375$\pm$0.139 & 0.699$\pm$0.056 & 0.624$\pm$0.107 & 0.723$\pm$0.052 & 0.649$\pm$0.106 \\
jnk3 & 0.065$\pm$0.034 & 0.032$\pm$0.017 & 0.108$\pm$0.017 & 0.054$\pm$0.005 & 0.110$\pm$0.017 & 0.055$\pm$0.005 \\
median1 & 0.245$\pm$0.026 & 0.205$\pm$0.024 & 0.304$\pm$0.017 & 0.269$\pm$0.007 & 0.310$\pm$0.018 & 0.276$\pm$0.006 \\
median2 & 0.200$\pm$0.013 & 0.185$\pm$0.015 & 0.234$\pm$0.019 & 0.226$\pm$0.020 & 0.237$\pm$0.019 & 0.230$\pm$0.021 \\
mestranol\_similarity& 0.433$\pm$0.060 & 0.383$\pm$0.059 & 0.508$\pm$0.064 & 0.484$\pm$0.064 & 0.516$\pm$0.065 & 0.494$\pm$0.065 \\
osimertinib\_mpo& 0.725$\pm$0.032 & 0.665$\pm$0.029 & 0.784$\pm$0.044 & 0.769$\pm$0.042 & 0.790$\pm$0.045 & 0.780$\pm$0.043 \\
perindopril\_mpo& 0.391$\pm$0.006 & 0.355$\pm$0.007 & 0.454$\pm$0.027 & 0.440$\pm$0.020 & 0.460$\pm$0.029 & 0.448$\pm$0.022 \\
qed & 0.916$\pm$0.025 & 0.887$\pm$0.028 & 0.938$\pm$0.012 & 0.932$\pm$0.016 & 0.940$\pm$0.011 & 0.937$\pm$0.015 \\
ranolazine\_mpo& 0.600$\pm$0.077 & 0.541$\pm$0.071 & 0.709$\pm$0.076 & 0.690$\pm$0.076 & 0.719$\pm$0.076 & 0.705$\pm$0.077 \\
scaffold\_hop& 0.485$\pm$0.005 & 0.469$\pm$0.005 & 0.513$\pm$0.013 & 0.505$\pm$0.011 & 0.515$\pm$0.014 & 0.508$\pm$0.012 \\
sitagliptin\_mpo& 0.219$\pm$0.080 & 0.141$\pm$0.105 & 0.299$\pm$0.083 & 0.215$\pm$0.156 & 0.307$\pm$0.084 & 0.222$\pm$0.161 \\
thiothixene\_rediscovery& 0.336$\pm$0.033 & 0.307$\pm$0.028 & 0.428$\pm$0.014 & 0.403$\pm$0.029 & 0.437$\pm$0.012 & 0.412$\pm$0.029 \\
troglitazone\_rediscovery& 0.256$\pm$0.071 & 0.230$\pm$0.064 & 0.305$\pm$0.085 & 0.289$\pm$0.091 & 0.310$\pm$0.087 & 0.295$\pm$0.094 \\
valsartan\_smarts& 0.000$\pm$0.000 & 0.000$\pm$0.000 & 0.000$\pm$0.000 & 0.000$\pm$0.000 & 0.000$\pm$0.000 & 0.000$\pm$0.000 \\
zaleplon\_mpo& 0.396$\pm$0.049 & 0.327$\pm$0.047 & 0.484$\pm$0.037 & 0.447$\pm$0.028 & 0.492$\pm$0.036 & 0.459$\pm$0.026 \\
\midrule
Sum & 9.788 & 8.669 & 11.862 & 11.161 & 12.067 & 11.410
\end{tabular}
      \end{sc}
    \end{small}
  \end{center}
  \vskip -0.1in
\end{table*}

\begin{figure}
    \centering
    \includegraphics[width=1\linewidth]{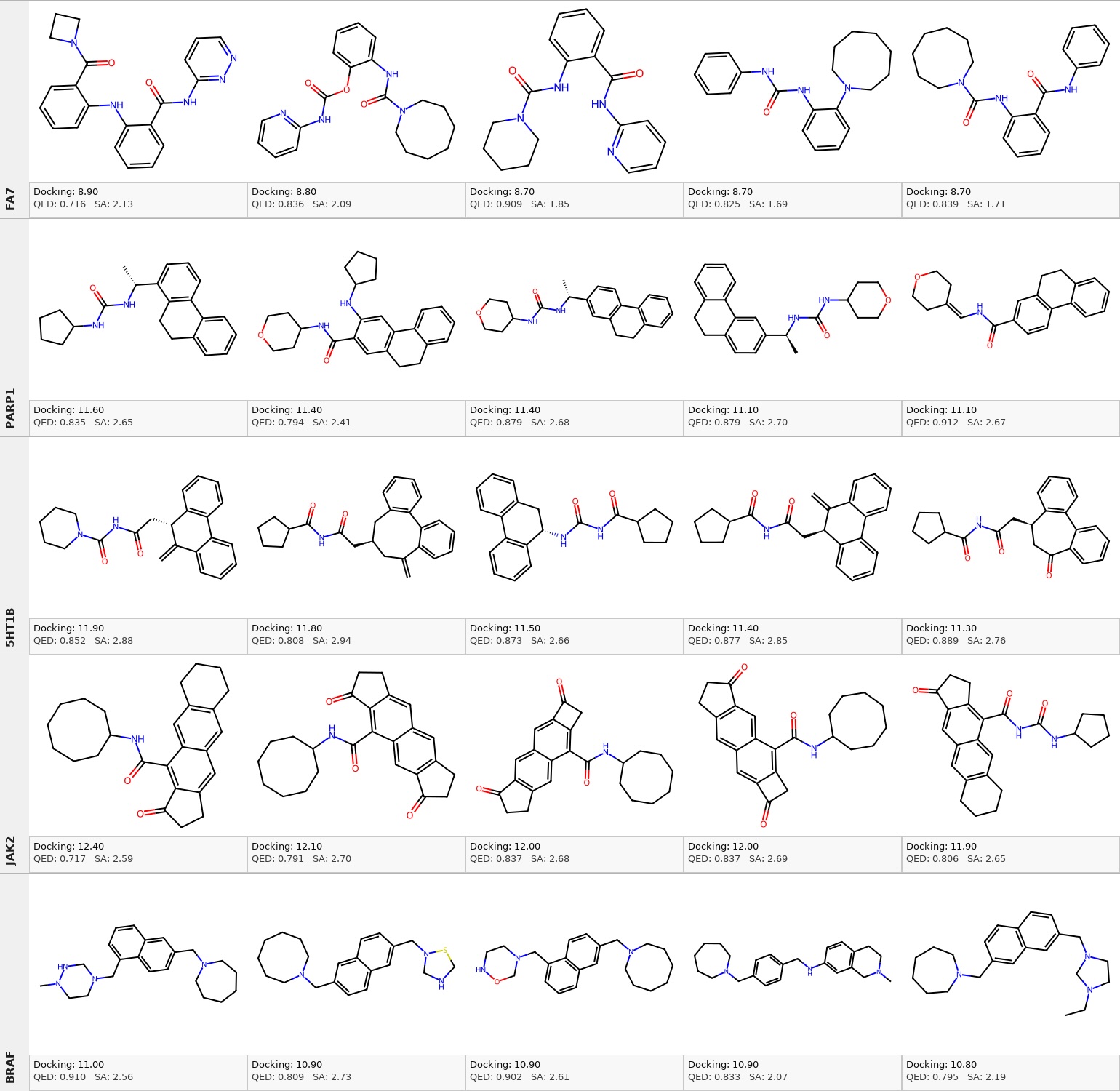}
    \caption{\textbf{Top 5 strict hits from one seed of SEGO for each target. }The top 5 docking scores from the strict hits from a single seed of SEGO run. They show local chemical space exploration with moderate diversity. This is to be expected and is similar to Saturn, overall diversity from pooled runs is higher.}
    \label{fig:samplemols}
\end{figure}


\end{document}